\documentclass[letterpaper, 10 pt, conference]{ieeeconf}  %

\IEEEoverridecommandlockouts                              %

\usepackage{graphicx}
\usepackage[font=small]{caption}
\usepackage{amsmath} %
\usepackage{amssymb}  %
\usepackage{bm} %
\usepackage[noend]{algorithmic}
\usepackage{algorithm}
\usepackage{xcolor}
\usepackage{booktabs}
\usepackage{siunitx}
\usepackage{tikz}
\usetikzlibrary{positioning}
\usepackage{hyperref}
\renewcommand{\vec}[1]{\bm{#1}}

\newcommand{\algfull}{Grasp-Optimized Motion Planning for Fast Inertial Transport}
\newcommand{\algname}{GOMP-FIT}

\newcommand{\remark}[3]{{\color[hsb]{#2,1.0,0.7} [#1: #3]}}
\renewcommand{\remark}[3]{}

\newcommand{\NEW}[1]{#1}

\title{\LARGE \bf
\algname{}: Grasp-Optimized Motion Planning \\for Fast~Inertial~Transport
}

\author{Jeffrey Ichnowski$^{1}$, Yahav Avigal$^{1}$, Yi Liu$^{1}$, and Ken Goldberg$^{1}$%
\thanks{$^{1}$AUTOLab,
        University of California, Berkeley, CA
        {\tt\small \{jeffi, yahav\_avigal, yiliu77, goldberg\}@berkeley.edu}}%
}

\begin{document}

\maketitle
\thispagestyle{empty}
\pagestyle{empty}

\begin{abstract}
High-speed motions in pick-and-place operations are critical to making robots cost-effective in many automation scenarios, from warehouses and manufacturing to hospitals and homes.  However, motions can be too fast---such as when the object being transported has an open-top, is fragile, or both.  One way to avoid spills or damage, is to move the arm slowly.  We propose an alternative: Grasp-Optimized Motion Planning for Fast Inertial Transport (GOMP-FIT), a time-optimizing motion planner based on our prior work, that includes constraints based on accelerations at the robot end-effector.  With GOMP-FIT, a robot can perform high-speed motions that avoid obstacles and use inertial forces to its advantage.  In experiments transporting open-top containers with varying tilt tolerances, whereas GOMP computes sub-second motions that spill up to 90\,\% of the contents during transport, GOMP-FIT generates motions that spill 0\,\% of contents while being slowed by as little as 0\,\% when there are few obstacles, 30\,\% when there are high obstacles and 45-degree tolerances, and 50\,\% when there 15-degree tolerances and few obstacles.  Videos and more at: \url{https://berkeleyautomation.github.io/gomp-fit/}.

\end{abstract}

\section{Introduction}
\label{sec:introduction}

Fast and reliable robot pick-and-place motion is increasingly a bottleneck in automated warehouses.  Motions that are too slow, while usually safe, reduce robot throughput, while motions that are too fast can be unreliable, leading to dropped objects due to shearing forces between the object and the gripper that holds it. Our prior work, Grasp-Optimized Motion Planning (GOMP~\cite{ichnowski2020gomp}), showed that simultaneously optimizing grasp pose and pick-and-place motion could allow for rapid object transport.  However, given the torque available to industrial robot arms, the high-speed motions that GOMP generates could lead to product damage and spills.  In this work, we propose a \emph{fast inertial transport} (FIT) problem and algorithm to address it,
in which a robot transports objects at high speeds, banking and limiting motions against inertial forces to safely and reliably place the object.

\begin{figure}[t]
    \centering
    \includegraphics[height=1.31in]{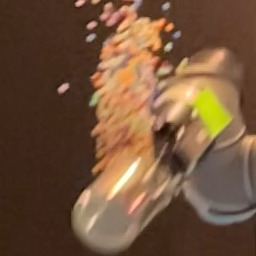}\hfill%
    \includegraphics[height=1.31in]{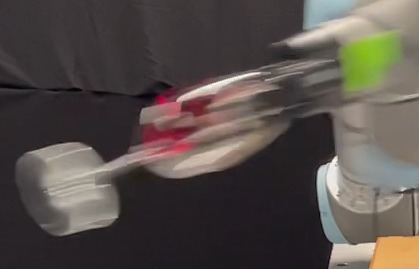}%
    \\[0.05in]
    \includegraphics[width=3.4in]{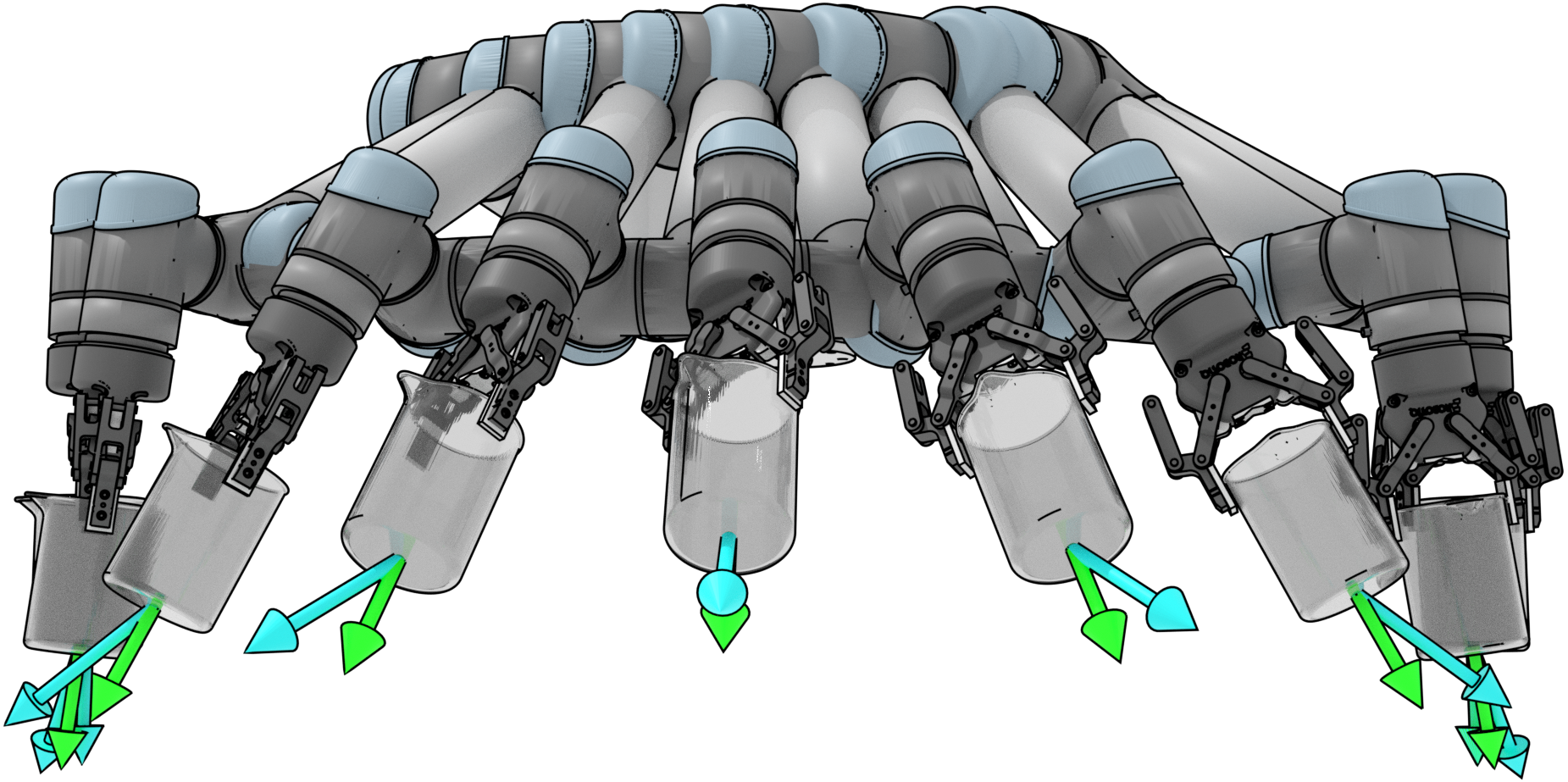}
    \caption{\textbf{High-speed transport-aware motion planning}
    \textbf{Top left:} without considering inertial effects, fast transport can lead to spills and damage.
    \textbf{Top right:} \algname{} computes high-speed transport motions for objects such as a filled wineglass, without spilling or breaking fragile objects. 
    \textbf{Bottom:} %
    \algname{} constrains the alignment between the end-effector accelerations due to inertial forces and gravity (\textcolor{cyan!75!black}{cyan arrow}), against the normal defined by the container (\textcolor{green!75!black}{green arrow}).  Here it is limited to 30$^\circ$.  The optimization rotates the robot's joints to keep within the constraint.
    }
    \label{fig:beads}\label{fig:motion}
\end{figure}

To enable FIT, we propose \emph{\algfull{}} (\algname{}) which computes time-optimized motions while taking into account end-effector and object acceleration constraints imposed by the object being transported.
\algname{} incorporates constraints for open-top container and fragile object transport, combinations thereof, and potentially more.
For open-top containers, \algname{} aligns inertial accelerations so that the contents remain in the container.
For fragile-object transport, \algname{} limits the magnitude of accelerations of the transported object to avoid exceeding a shock threshold.
For fragile open-top containers, \algname{} constrains both magnitude and alignment of accelerations.
As most industrial robot arms do not provide direct access to motor torques, \algname{} instead operates in the joint configuration space, relying on the robot's black-box controller to follow trajectories.

In experiments, a physical UR5 robot transports various objects with end-effector acceleration requirements, including: transporting cups filled to various levels with beads without spilling their contents, transporting an IMU to measure accelerations that a fragile object would experience in transit, and moving a filled wineglass.  Experiments suggest that \algname{} can achieve near 100\,\% success on these tasks while slowing as little as 0\,\% when there are few obstacles, 30\,\% when there are high obstacles and 45-degree tolerances, and 50\,\% when there are 15-degree tolerances and few obstacles compared to GOMP.
This work contributes:
\begin{enumerate}
    \item a formulation of the fast inertial transport motion planning problem,
    \item a sequential convex program, including non-convex constraints on accelerations at the end-effector, and a sequential quadratic program to solve it,
    \item data from experiments on a physical UR5 robot transporting cups filled with beads, a fragile object and a filled wineglass.
\end{enumerate}

\section{Related Work}

\subsection{Motion Planning and Optimization}
Motion planning seeks to produce paths from start to goal while avoiding obstacles.
Sampling-based motion planners such as PRM~\cite{Kavraki1996_TRA}, RRG~\cite{Karaman2011_IJRR}, and RRT~\cite{LaValle2000_WAFR} have favorable properties such as probabilistic completeness and asymptotic optimality, but they may be slow to converge. %
Such planners also typically require a post-processing  step to remove redundant or jerky motion.
Optimization-based motion planners such as TrajOpt~\cite{schulman2013finding}, STOMP~\cite{kalakrishnan2011stomp}, CHOMP~\cite{ratliff2009chomp}, and KOMO~\cite{toussaint2014newton} can produce smooth trajectories while fulfilling a set of constraints such as obstacle avoidance through interior point optimization, stochastic gradients, and covariant gradient descent. In prior formulations, the trajectory is discretized into a series of waypoints with a minimization objective such as sum of squares velocity. %
Motion planners such as GOMP~\cite{ichnowski2020gomp} and DJ-GOMP~\cite{ichnowski2020djgomp} add dynamics constraints and shrinking horizon lengths to find a time-optimized trajectory, and apply jerk limits to avoid damaging the robot. Unlike previous work, this paper applies additional constraints to the end-effector acceleration to prevent an object held by a robot arm from damage or detachment in high-acceleration trajectories, something that constraints on configurations and their derivatives alone cannot do.

\subsection{End-Effector Constraints}

Placing constraints on the end-effector motion path is a requisite part of many problems, including fast inertial transport.  Yao and Gupta~\cite{yao2007path} address the path-planning problem with general end-effector constraints by exploring the task space for feasible end-effector poses through a sampling-based planner.  Li et al.~\cite{li2021constrained} propose using RL to compute motion plans for dual-arm manipulators with floating base in space.  They propose to take into account a velocity constraint of the end-effector in the planning process to improve the robustness of the algorithm.  But these methods do not support acceleration constraints for the fast inertial transport problem we propose.

\subsection{Planning for Dynamic Manipulation}

Dynamic manipulation research uses dynamic properties such as momentum to achieve a task---whether prehensile or non-prehensile.  Lynch and Mason~\cite{lynch1996dynamic} exploit centrifugal and Coriolis forces to manipulate objects using low-degree-of-freedom robots.
Lynch and Mason~\cite{lynch1999dynamic} also formulate an SQP for dynamic non-prehensile manipulation of objects, such as snatching, throwing, and rolling.  They integrate constraints based on a 2D formulation of the problem but do not consider obstacles.
Srinivasa et al.~\cite{srinivasa2005using} propose employing constraints on accelerations at the end point to perform dynamic non-prehensile rotation of an object.
Kim et al.~\cite{kim2014catching} propose a method to rapidly computes motions to catch an object in flight.
Mucchiani and Yim~\cite{mucchiani2021dynamic} propose a method to use inertial effects to dynamically sweep up an object and stabilize it using a passive end-effector.
In contrast, we propose a 3D prehensile planner for safe and reliable object transport around obstacles.

A promising line of research, especially for objects with unknown or difficult-to-model dynamics, is to employ learning.
Zeng et al.~\cite{zeng2020tossingbot} propose TossingBot that learns parameters of a pre-defined dynamic motion to throw objects into target bins.
Zhang et al.~\cite{zhang2021rotla} propose a method that learns a sweeping dynamic motion for ropes to hit targets, weave through obstacles, or knock objects down.
Wang et al.~\cite{wang2020swingbot} propose SwingBot, that uses inhand tactile feedback to learn how to swing up previously unseen objects.  These methods rely on a learned model for dynamic manipulation, while we propose using Newtonian physics.

Another promising approach is to combine dynamics considerations and sampling-based planning.
Pham et al.~\cite{pham2013kinodynamic,pham2014planning} proposed admissible velocity propagation (AVP) and a method to perform kinodynamic planning in a reduced dimensionality state space, and Lertkultanon and Pham~\cite{lertkultanon2014dynamic} formulate a ZMP constraint for AVP and integrate bi-directional RRT to non-prehensile object transportation.  Unlike the AVP formulation, \algname{} integrates all constraints, including collision, in a single optimization.

To enable real-world interaction between robots and the environments they touch, researchers have looked into optimizing motions with contacts.
Posa and Tedrake~\cite{posa2013direct} and Posa et al.~\cite{posa2014direct} propose simultaneously optimizing trajectories and contact points.
Hauser~\cite{hauser2014fast} proposes a trajectory optimization that considers contact forces and convex time scaling.
Luo and Hauser~\cite{luo2017robust} propose integrating feedback and learning to integrate confidence into the optimization, and apply it to the Waiter's Problem of stably transporting objects on a moving platform.
While these methods consider contacts, they do not consider obstacle avoidance.

\subsection{Dynamic Object Transport}

Most closely related to this paper is research into transport of objects that takes into account inertial effects.
Bernheisel and Lynch~\cite{bernheisel2004stable} propose a Waiter's Problem in which object assemblies are stably transported subject to inertial and gravity forces.  They focus on multi-part assemblies and assume quasistatic motion in which only the velocity direction matters.  In contrast we focus on a containment and acceleration-based inertial effects.
Wan et al.~\cite{wan2020waiter} demonstrates beverage transport with %
jerk limits help prevent spills during transport. In contrast, we show that jerk limits alone will not prevent spills when using a manipulator arms.
Acharya et al.~\cite{acharya2020nonprehensile} 
propose methods to apply minimum-time s-curve trajectories to the Waiter's Problem.
To transport objects with unknown mass parameters, Lee and Kim~\cite{lee2017constraint} perform online estimation of payload parameters to integrate into the trajectory generation for cooperative aerial manipulators.
For fast transport of objects held in a suction grasp, Pham and Pham~\cite{pham2019critically} propose combining RRT with a trajectory time parameterizer that remains within the suction contact stability constraint.
In contrast, we propose a single optimization process that incorporates all necessary constraints to perform fast inertial transport.

\NEW{%
While the primary intent for GOMP-FIT is not liquid transport, in experiments we show spill-free transport of a liquid.  The related problem of slosh-free transport requires constraining accelerations of a container.  
Chen et al.~\cite{chen2007using} add liquid transport to the Waiter's problem and use an acceleration filter to orient the end-effector to counter sloshing effects through Cartesian control.
Aribowo et al.~\cite{aribowo2015integrated} decouple liquid transport into two steps: computing a translation trajectory, then input shaping to counter sloshing by rotating the end-effector. %
Yano et al.~\cite{yano2001robust}, Reyhanoglu et al.~\cite{reyhanoglu2013nonlinear}, Consolini et al.~\cite{consolini2013minimum} and Moriello et al.~\cite{moriello2017control} propose using feedback and feed-forward control to avoid sloshing under varying assumptions such as linear actuation and absence of obstacles.  In contrast to these works, GOMP-FIT integrates end-effector acceleration constraints into a single time-optimizing method. %
}

\section{Problem Statement}
\label{sec:problem_statement}

Let $\vec{q} \in \mathcal{C}$ be the complete specification of a robot's degrees of freedom, where $\mathcal{C}$ is the space of all configurations.
Let $\mathcal{C}_\mathrm{obs} \subset \mathcal{C}$ be the set of configurations that are in an obstacle, and $\mathcal{C}_\mathrm{free} = \mathcal{C} \setminus \mathcal{C}_\mathrm{obs}$ be the set of configurations that are not in collision.
Let $f_k:\mathcal{C} \rightarrow SO(3)$ be the forward kinematics function that computes the pose of the object in the robot's end-effector.  Let $f^{-1}_k : SE(3) \rightarrow \mathcal{C}$ be an inverse kinematic (IK) function that computes the robot's configuration, given a desired location of the end-effector.  The IK function may not always have a solution, and may not have a unique solution.
Let $\vec{x}^T = \begin{bmatrix} \vec{q}^T & \dot{\vec{q}}^T & \ddot{\vec{q}}^T \end{bmatrix} \in \mathcal{X}$ be the dynamic state of the robot at any moment in time, including the first and second derivative of the robot's configuration.
Let $f_a : \mathcal{X} \rightarrow \mathbb{R}^3$ be the linear acceleration at the end-effector including gravity and the inertial (fictitious) forces: Euler, Coriolis, and centrifugal.

Given a starting grasp $\vec{g}_\mathrm{pick} \in SE(3)$ and a placement pose $\vec{g}_\mathrm{place} \in SE(3)$, and constraints on the end-effector accelerations, the objective of \algname{} is to compute a trajectory $\tau : [0,T] \rightarrow \mathcal{C}$, where $T$ is the duration, $\tau(t) \in \mathcal{C}_\mathrm{free}$, $f(\tau(0)) = \vec{g}_\mathrm{pick}$, $f(\tau(T)) = \vec{g}_\mathrm{place}$, and the end-effector acceleration constraints are met at all $\tau(t)$.  Furthermore, the objective is to minimize the total trajectory time, subject to the robot's actuation limits.  Thus,
\begin{align*}
    \arg\min_\tau \quad & T(\tau) \\
    s.t. \quad & f_k(\tau(0)) = \vec{g}_\mathrm{pick}, \,  f_k(\tau(T)) = \vec{g}_\mathrm{place} \\
    & \tau(t) \in \mathcal{C}_\mathrm{free} \quad \forall t \in [0, T] \\
    & \tau(t), \dot\tau(t), \ddot\tau(t) \in \text{joint limits} \quad \forall t \in [0, T] \\
    & f_a(\tau(t)) \in \mathcal{A}
\end{align*}
where $T(\tau)$ is the trajectory's duration in seconds (and referenced without parameters equivalently), and $\mathcal{A}$ is the set of problem-specific end-effector acceleration constraints, such as keeping open-top container contents or avoiding excessive shock. 

\section{Method}
\label{sec:method}

To compute a high-speed motion that remains within end-effector acceleration constraints, we first discretize the problem, then formulate a non-convex optimization, and finally solve the optimization using a sequence of sequential quadratic programs based on TrajOpt~\cite{schulman2013finding} and GOMP~\cite{ichnowski2020gomp}.

\subsection{GOMP Background}

\algname{} is built on Grasp-Optimized Motion Planning (GOMP).  GOMP computes a time-optimized obstacle-avoiding trajectory that incorporates a degree of freedom around pick and place points.  This section reviews the GOMP formulation and optimization process.

\subsubsection{Trajectory Discretization}
To facilitate solving the \algname{} optimization problem, we first formulate a discretization of the trajectory.  The discretization serves a second purpose---all industrial robots operate with a fixed control frequency, and can follow trajectories at only that rate or an integer multiple of it.  %

GOMP and \algname{} first define a fixed number of waypoints ${0, 1, \ldots, H}$, each separated by a fixed time step $t_\mathrm{step}$.  The time step should be an integer multiple of the robot's control frequency, and $H$ should be sufficient to allow the problem to be solved (more details in Sec.~\ref{sec:method:time_opt}).  Each waypoint $\vec{x}_t$ includes a configuration $\vec{q}_t$ and first and second derivatives $\dot{\vec{q}}_t$ and acceleration $\ddot{\vec{q}}_t$.

To approximate $\dot{\vec{q}}_t$ and $\ddot{\vec{q}}_t$, we have the optimization process enforce a \emph{dynamics} constraint between each waypoint in the following form:
\begin{align}
    \vec{q}_{t+1} & = \vec{q}_t + \dot{\vec{q}}_t t_\mathrm{step} + \left(\frac 13 \ddot{\vec{q}}_t + \frac 16 \ddot{\vec{q}}_{t+1}\right) t_\mathrm{step}^2 \label{eqn:config_dynamics} \\
    \dot{\vec{q}}_{t+1} & = \dot{\vec{q}}_t + \frac 12 \left( \ddot{\vec{q}}_t + \ddot{\vec{q}}_{t+1} \right) t_\mathrm{step}. \label{eqn:velocity_dynamics}
\end{align}
This discretization comes from integrating a linear jerk between waypoints.

\subsubsection{Sequential Convex Optimization}

To solve the discretized problem, we formulate and solve a sequential quadratic program of the following form:
\begin{align*}
    \min_{\vec{x}} \quad & \frac 12 \vec{x}^T P \vec{x} + \vec{p}^T \vec{x} \\
    \text{s.t.} \quad & A \vec{x} \le \vec{b},
\end{align*}
where $P$ is a positive semi-definite matrix for the quadratic costs, $\vec{p}$ is a linear cost vector, and the matrix $A$ and vector $\vec{b}$ define the linear constraints.

We construct $A$ and $\vec{b}$ to include the \emph{convex} constraints:
(a) the dynamics from equations~\ref{eqn:config_dynamics} and \ref{eqn:velocity_dynamics}; 
(b) the actuation limits, e.g., box bounds of configuration, velocity, and acceleration; and
(c) jerk limits from the finite difference of accelerations.

To handle \emph{non-convex} constraints from
(i) grasp and placement configurations and degrees of freedom;
(ii) obstacle avoidance; and
(iii) end-effector acceleration limits, 
we add linearizations of the constraints to $A$ and $\vec{b}$ with slack variables constrained to be positive.

As with GOMP, we establish the trust region around the configurations of the waypoints, and leave the other optimization variables (e.g., $\dot{\vec{q}}_t$ and $\ddot{\vec{q}}_t$) otherwise untouched.

\subsubsection{Linearization of Non-Convex Constraints}

To enable obstacle avoidance, grasp-optimization, and now acceleration constraints, GOMP and \algname{} formulates a (non-convex) constraint of the form $g(\vec{\vec{x}}) \le c$, and 
linearizes the function around the current iterate $\vec{x}^{(k)}$ as:
\[
J_g \vec{x}^{(k+1)} \le J_g \vec{x}^{(k)} - g(\vec{x}^{(k)}) - c,
\]
where $J_g$ is the Jacobian evaluated at $\vec{x}^{(k)}$, and $\vec{x}^{(k+1)}$ is the optimization variable for the next iteration.  These coefficients are then added to $A$ and $\vec{b}$. %

\subsubsection{Time Optimization}
\label{sec:method:time_opt}

To minimize trajectory time, GOMP and \algname{} repeatedly solve the optimization with a shrinking horizon $H$ until the SQP returns failure, and uses trajectory from the minimum horizon that succeeded.

\subsection{\algname{} End-Effector Acceleration Constraints}

\begin{figure}[t]
    \centering
    \input{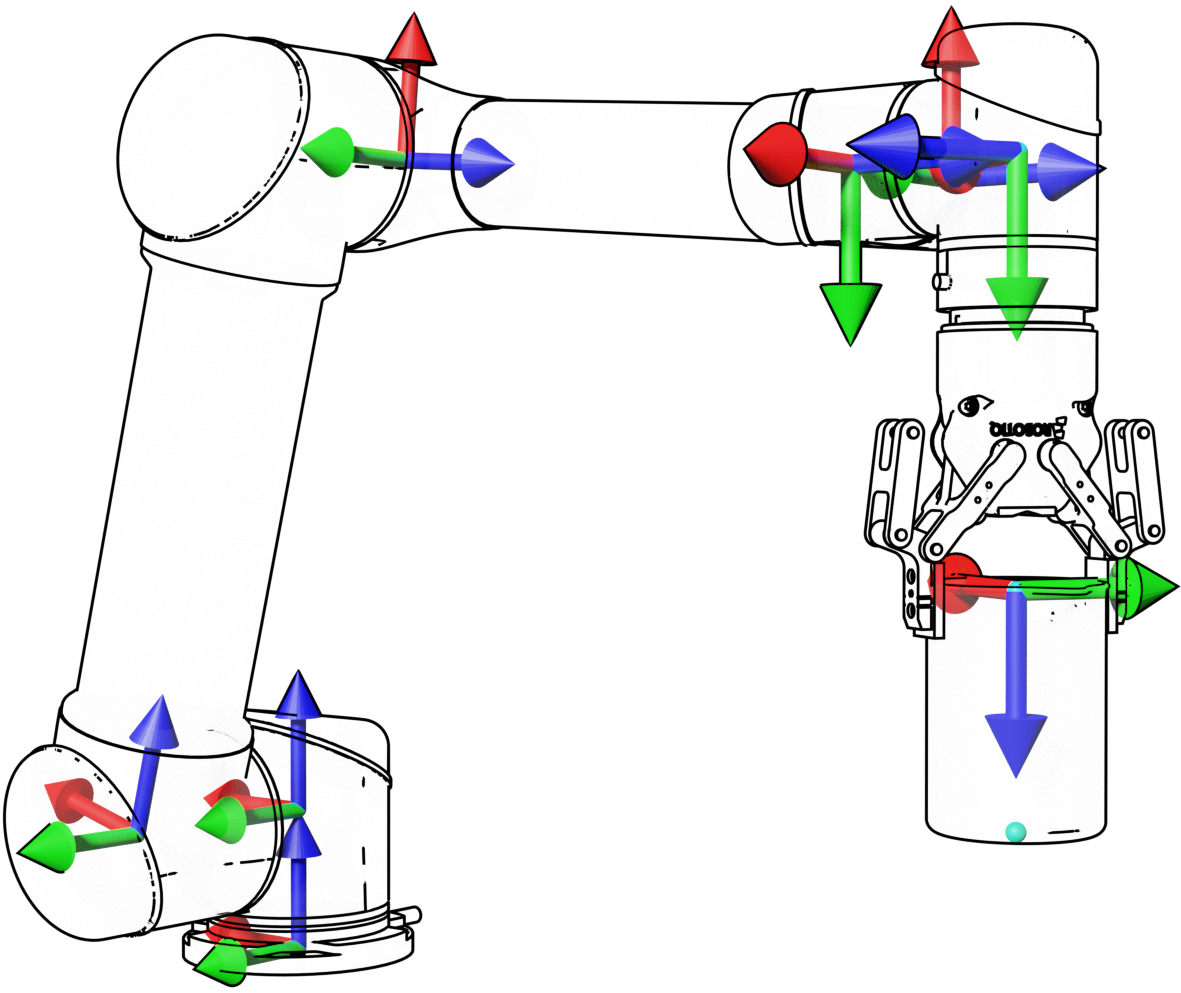}
    \caption{\textbf{Frames and parameters used in \algname{} computation}.  Each set of arrows is a coordinate frame associated with a joint.  The recursive Newton-Euler algorithms computes angular velocities and their derivatives, and linear accelerations at each frame $R$. The translation between frames $r$.  At the end-effector, $f_n(\vec{q})$ is the normal we align to inertial forces.  %
    }
    \label{fig:frames}
\end{figure}

To constrain inertial effects, we first compute the acceleration at the end-point due to gravity, Euler, Coriolis, and centrifugal forces; and then form the constraint. %
To compute the effect of all joint motions on the end-effector acceleration, we employ the forward pass of the Recursive Newton Euler (RNE) method~\cite{luh1980line}, summarized in Alg.~\ref{alg:rne}.  RNE requires the rotation ${}^{i-1}R_i$ between joints $i-1$ and $i$; the angular velocity vector $\vec{\omega}$ and its derivative $\dot{\vec{\omega}}$; and the translation between successive joints $\vec{r}_i$.  We do not use the backward pass of the RNE, as the forward pass is sufficient. %

We compute the acceleration at the end-effector $\vec{a}_\mathrm{ee} = \vec{g} - \mathrm{RNE}(\vec{q}, \dot{\vec{q}}, \ddot{\vec{q}})$, where $\vec{g}$ is the gravity vector.
We define $f_n(\vec{q}) : \mathcal{C} \rightarrow \mathbb{R}^3$ as the forward kinematic to the grasp vector (Fig.~\ref{fig:motion})---i.e., normal of a container's surface.
We then formulate the non-convex constraints for object transport.  For notation convenience, we omit the subscript $t$, but these constraints apply to all waypoints.

\begin{algorithm}[t]
   \caption{Recursive Newton Euler}
   \label{alg:rne}
\begin{algorithmic}[1]
    \STATE \textbf{Input:} Configuration and its derivatives $\vec{q}, \dot{\vec{q}}, \ddot{\vec{q}} $
    \STATE /* Notation: $\vec{q}_{[\mathrm{i}]} \in \mathbb{R}$ is $\mathrm{i}$-th joint angle */
    \STATE $\vec{\omega}_0, \dot{\vec{\omega}}_0, \vec{a}_0 \leftarrow 0, 0, 0$
    \FOR{$\mathrm{i} = 1, 2, \ldots, \text{number of joints}$}
        \STATE $\vec{\omega}_\mathrm{i} \leftarrow {}^{\mathrm{i}-1}R_\mathrm{i}^T (\vec{\omega}_{\mathrm{i}-1} + \dot{\vec{q}}_{[\mathrm{i}]} z_\mathrm{i})$
        \STATE $\dot{\vec{\omega}}_\mathrm{i} \leftarrow {}^{\mathrm{i}-1}R_\mathrm{i}^T (\dot{\vec{\omega}}_{\mathrm{i}-1} + \ddot{\vec{q}}_{[\mathrm{i}]} z_\mathrm{i} + \dot{\vec{q}}_{[\mathrm{i}]} \vec{\omega}_{\mathrm{i}-1} \times z_\mathrm{i})$
        \STATE $\vec{a}_\mathrm{i} \leftarrow {}^{\mathrm{i}-1}R_\mathrm{i}^T \vec{a}_{\mathrm{i}-1} + \dot{\vec{\omega}}_\mathrm{i} \times \vec{r}_{\mathrm{i}} + \vec{\omega}_\mathrm{i} \times (\vec{\omega}_\mathrm{i} \times \vec{r}_\mathrm{i})$
    \ENDFOR
    \RETURN $^{0}R_N \vec{a}_N$
\end{algorithmic}
\end{algorithm}

\subsubsection{Open-Top Containers}
To transport open-top containers, we constrain trajectories to keep the acceleration at the end-effector within a user-defined threshold angle $\theta_\mathrm{max}$ of the container normal (Fig.~\ref{fig:motion}).  This constraint has the form:
\begin{equation}\label{eqn:alignment}
\cos^{-1} ( \ddot{\vec{a}}_{\mathrm{ee}} \cdot f_n(\vec{q}) / \lVert \ddot{\vec{a}}_\mathrm{ee} \rVert ) \le \theta_\mathrm{max}.
\end{equation}
This can equivalently be expressed as
\[
\vec{a}_{\mathrm{ee}} \cdot f_n(\vec{q}) / \lVert \vec{a}_\mathrm{ee} \rVert \ge \cos \theta_\mathrm{max}.
\]

\subsubsection{Fragile Objects}
When transporting fragile objects, we constrain trajectories to avoid exceeding a threshold acceleration $a_\mathrm{max}$ in any direction.  This constraint is:
\begin{equation}\label{eqn:fragile}
    \lVert \vec{a}_\mathrm{ee} \rVert \le a_\mathrm{max}.
\end{equation}

\subsubsection{Combining Constraints}

When transporting fragile objects in open-top containers, %
it is possible to include both Eqn.~(\ref{eqn:alignment}) and Eqn.~(\ref{eqn:fragile}) as independent constraints.

\subsection{Minimization Objective}

As the constraints on end-effector accelerations and their linearizations depend on the velocity and acceleration of the configuration, we find that adding a minimization objective based on the sum-of-squared velocities or accelerations can work against the linearizations.  These objectives tend to ``pull'' trajectories against the constraints.  We thus minimize the sum-of-squared jerks of the trajectory using a finite difference between waypoints:
\[
\frac 1{t^2_\mathrm{step}} \sum_{t=0}^{H-1} (\ddot{\vec{q}}_{t+1} - \ddot{\vec{q}}_t)^2.
\]
In practice, we find that assigning different weights to different joints benefits motion computation.  Specifically, we weigh joints closer to the base higher than joints closer to the end-effector, encouraging those joints to focus more on producing a smooth overall motion with minimal jerk whereas joints near the gripper are encouraged to fulfill other problem-specific constraints such as the start goal constraint.

\section{Experiments}
\label{sec:experiments}

\begin{figure*}
    \centering
    \begin{tabular}{@{}c@{\hspace{.05in}}c@{\hspace{.05in}}c@{\hspace{.05in}}c@{\hspace{.05in}}c@{}}
    \includegraphics[width=1.36in,clip,trim=425 0 300 0]{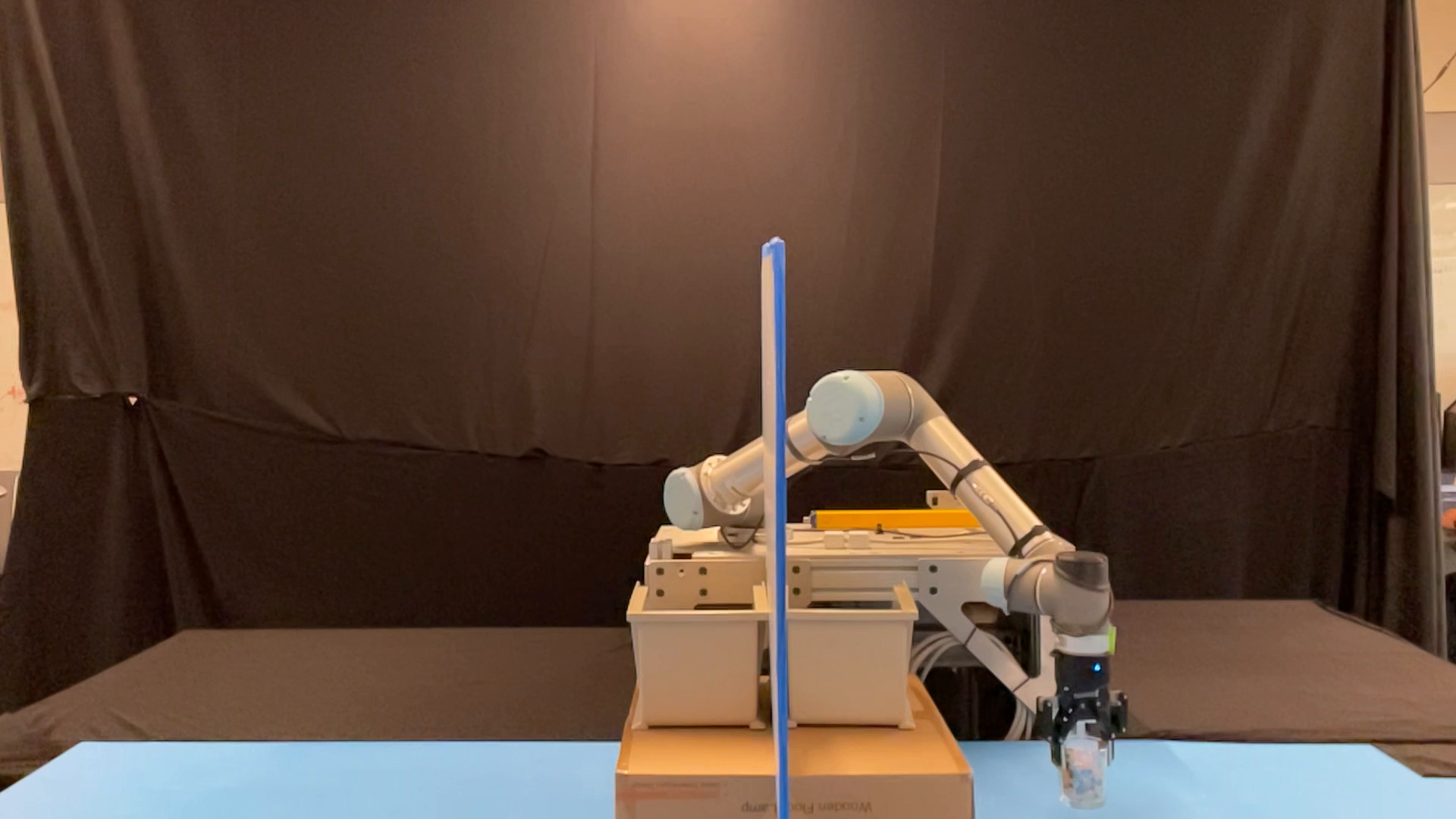} &
    \includegraphics[width=1.36in,clip,trim=425 0 300 0]{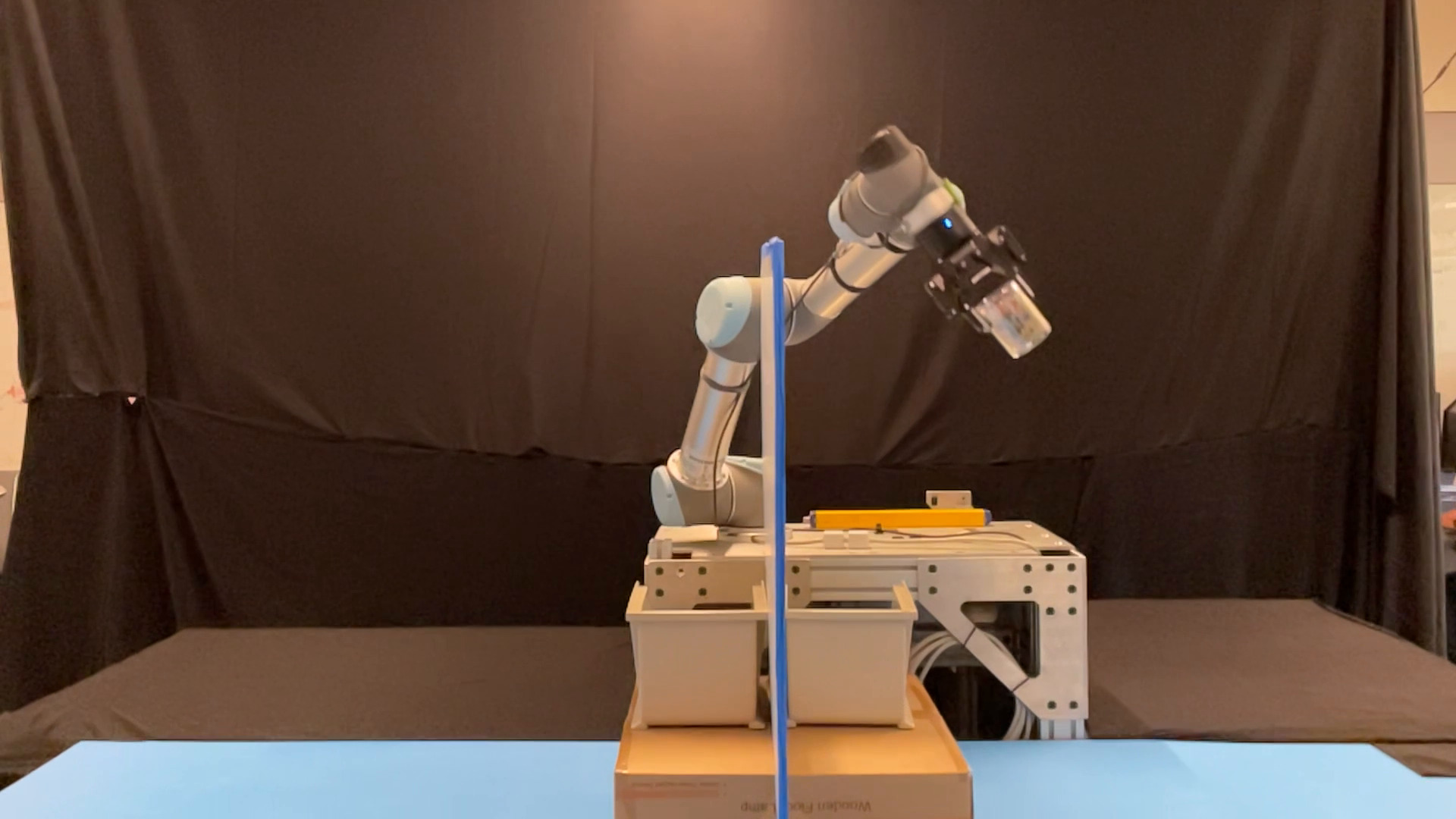} &
    \includegraphics[width=1.36in,clip,trim=425 0 300 0]{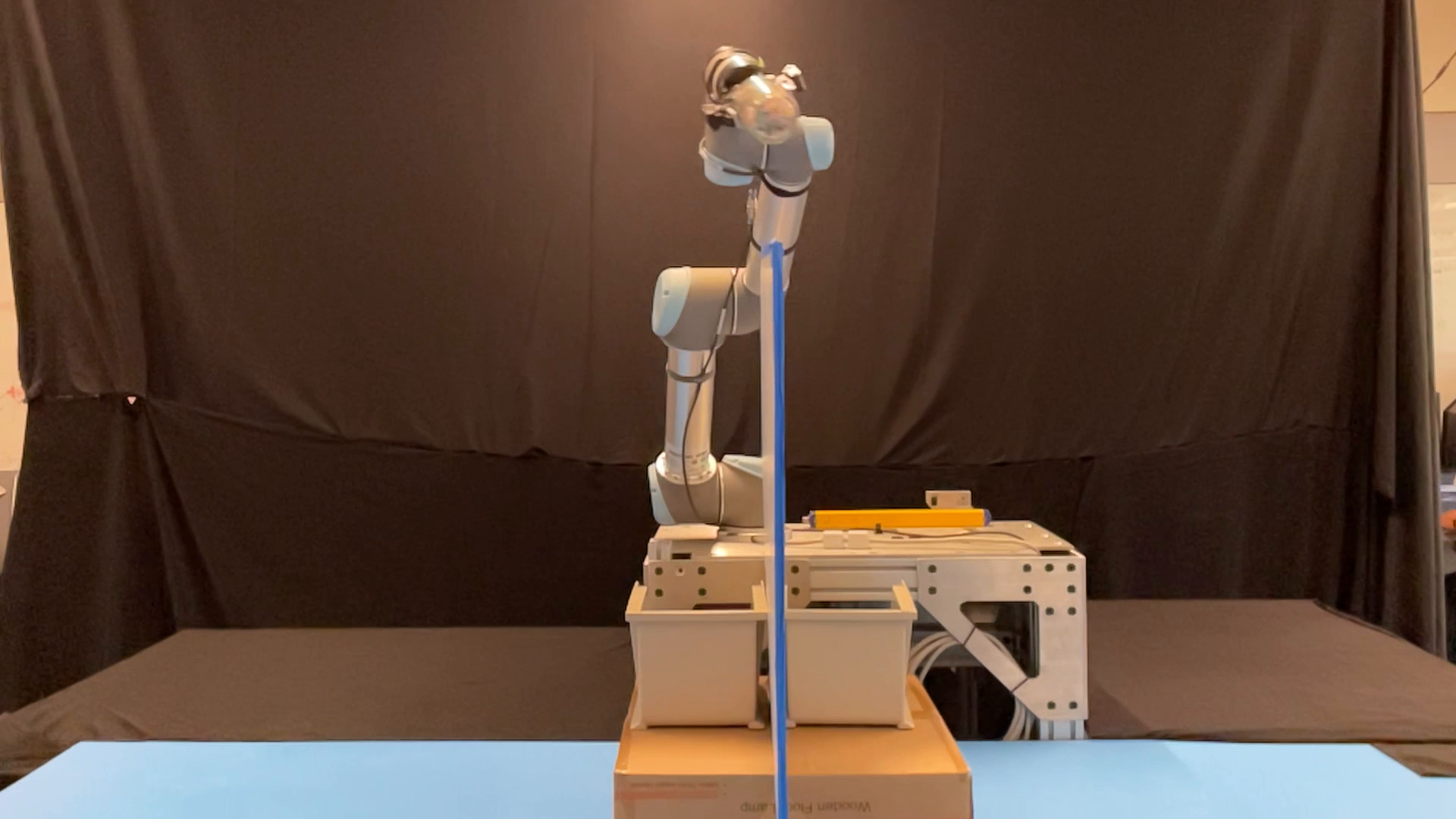} &
    \includegraphics[width=1.36in,clip,trim=425 0 300 0]{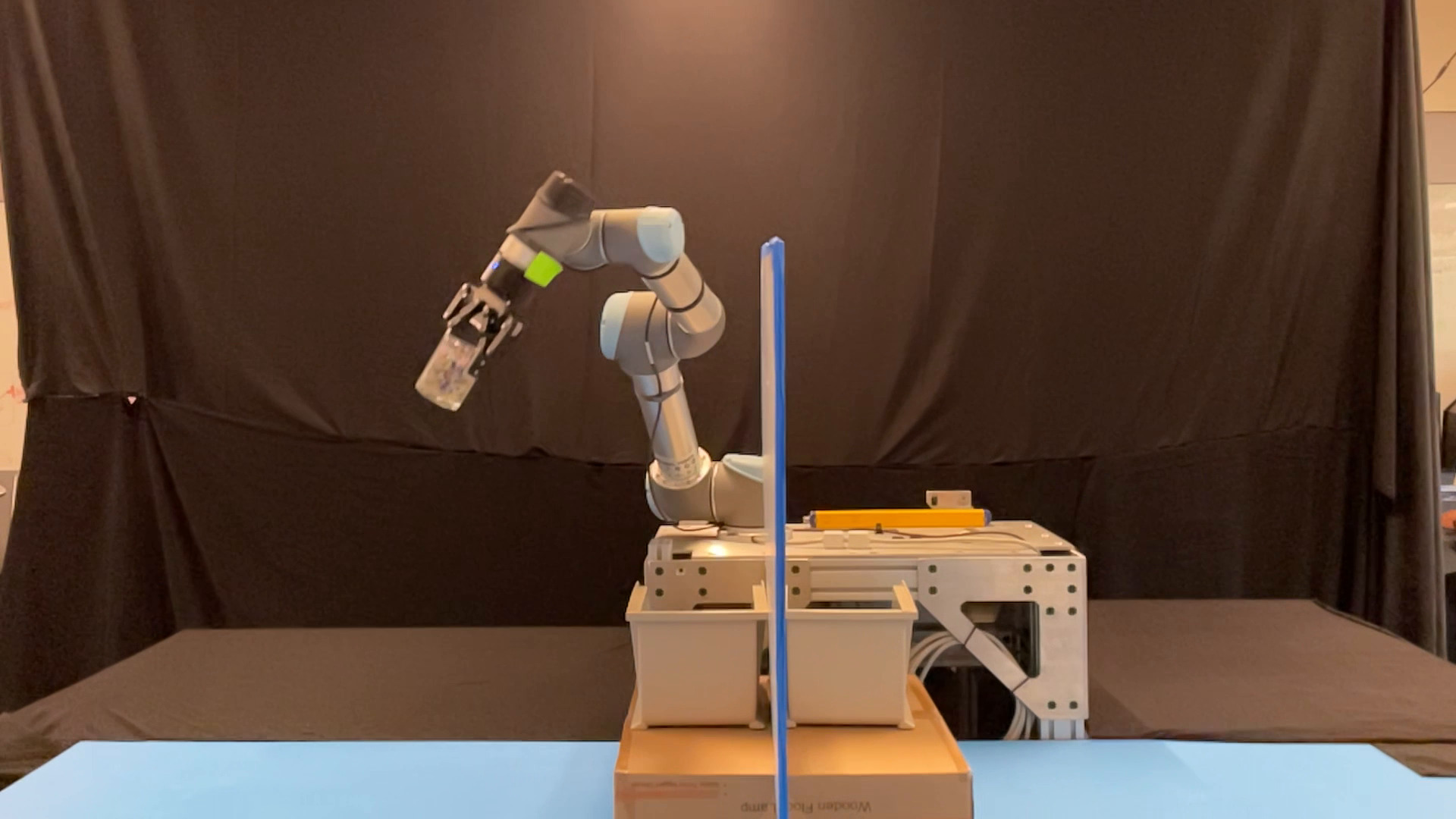} &
    \includegraphics[width=1.36in,clip,trim=425 0 300 0]{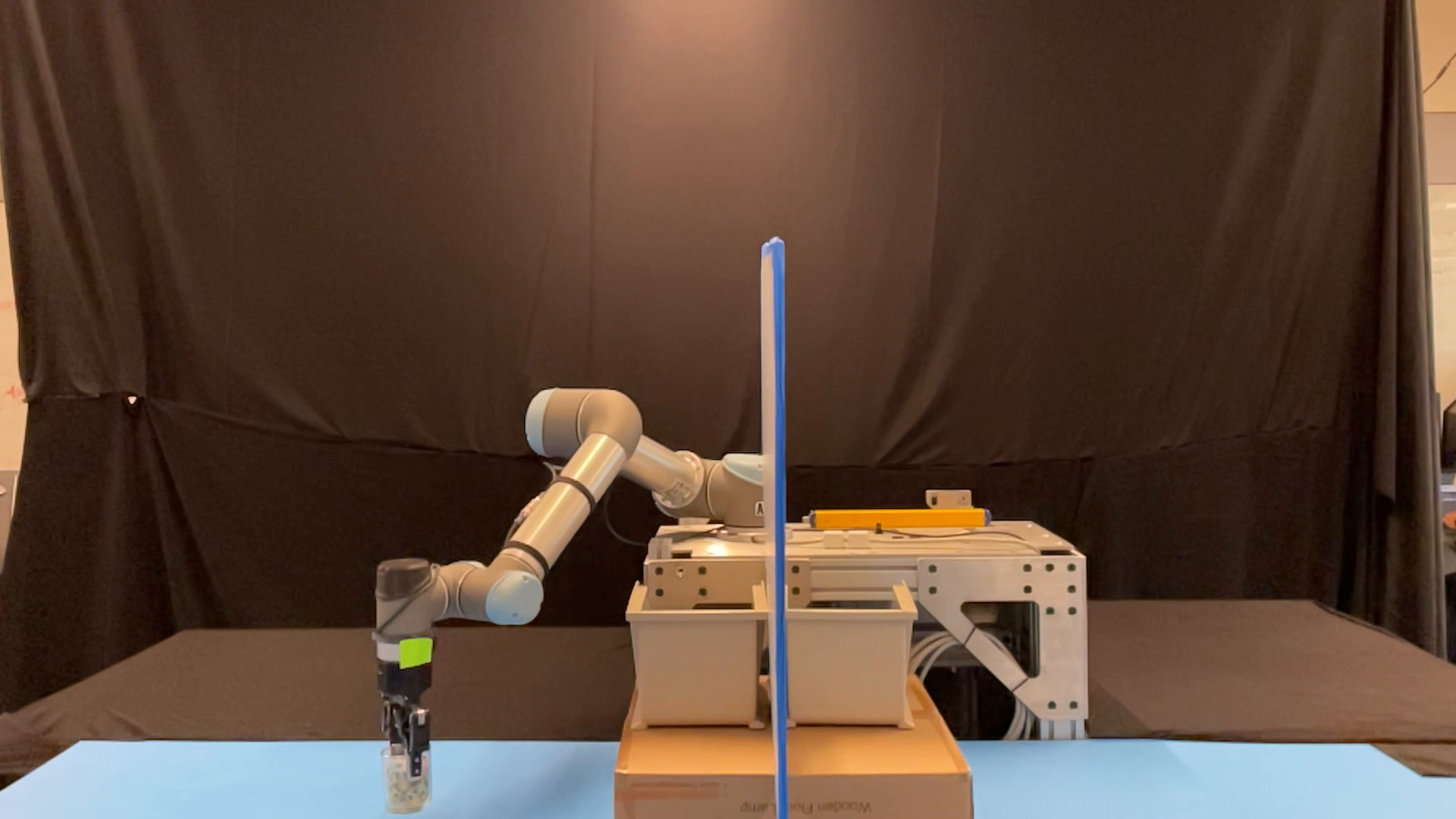} \\
    \includegraphics[width=1.36in]{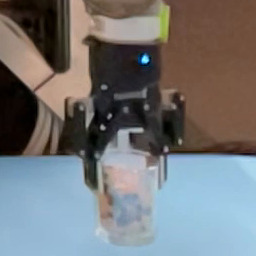} &
    \includegraphics[width=1.36in]{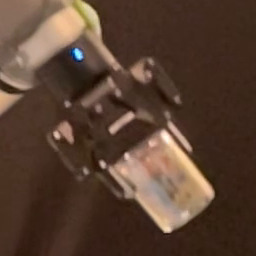} &
    \includegraphics[width=1.36in]{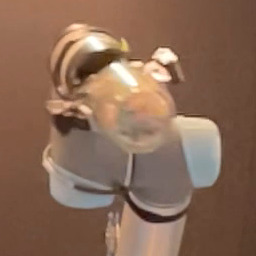} &
    \includegraphics[width=1.36in]{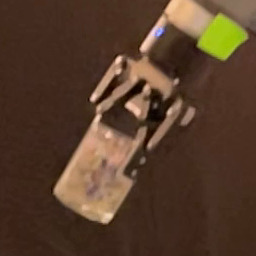} &
    \includegraphics[width=1.36in]{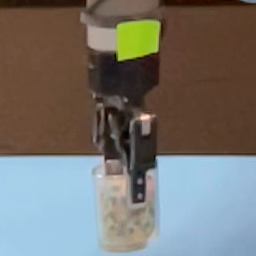} \\
    (a) & (b) & (c) & (d) & (e)
    \end{tabular}
    \caption{\textbf{Transporting an open-top container over a barrier.}
    Stills (\textbf{top}) and zoom-ins (\textbf{bottom}) over the course of a fast inertial transport motion.
    \algname{} computes a motion that transports an open-top container while avoiding a barrier 0.9\,m %
    above the floor, and keeping the end-effector accelerations aligned to a 45$^\circ$ to keep the contents in the container.  During this motion, the robot spills 0 beads.}
    \label{fig:over_the_top}
    \vspace{-12pt}
\end{figure*}

We experiment with \algname{} on a UR5 physical robot performing a series of tasks in which success is dependent on constraints on the end-effector acceleration. These tasks are:
(1) transporting an open-top container without spilling contents,
(2) transporting a fragile object without exceeding an acceleration threshold
(3) transporting wine held in a wine glass without spilling, or dropping the wineglass.

For baselines, we run both GOMP and J-GOMP.  This also forms an ablation in that \algname{} without the acceleration constraints is J-GOMP, and J-GOMP without the jerk limits and minimization is GOMP, as GOMP instead minimizes sum-of-squared accelerations at the joints.  Here, both GOMP and J-GOMP include a minor modification so that all baselines share the same dynamics constraints as \algname{}.  
Additionally, we add baselines of GOMP(+H) and J-GOMP(+H), which are GOMP and J-GOMP but with the same minimum horizon $H$ as computed by \algname{}.  These two baselines are to test whether it is just the slower speed of the trajectory, or if the additional constraints on end-effector accelerations are necessary for successful completion of the tasks.

\subsection{Open-Top Container Transport}

\begin{table}[t]
    \centering
    \scriptsize
    \begin{tabular}{@{}rrrrrr@{}}\toprule
         Tolerance & GOMP & J-GOMP & GOMP(+H) & J-GOMP(+H) & \algname{} \\
         \midrule
         45$^{\circ}$ & 89.7\% & 46.4\% & 84.5\% & 37.1\% & \textbf{0.0\%} \\
         30$^{\circ}$ & 90.0\% & 48.0\% & 47.0\% & 100.0\% & \textbf{0.0\%} \\
         20$^{\circ}$ & 90.3\% & 50.0\% & 39.4\% & 100.0\% & \textbf{0.0\%}\\
         15$^{\circ}$ & 91.2\% & 54.4\% & 41.2\% & 100.0\% & \textbf{0.0\%}\\
         \bottomrule
    \end{tabular}
    \caption{\textbf{Open-top container lost mass.}  We measure the percentage of mass lost when performing the transport of an open-top container using various methods.  The container starts with a consistent fill level for each tilt/tolerance angle.} %
    \label{tab:open_top_lost_mass}
\end{table}
\begin{table}[t]
    \centering
    \begin{tabular}{@{}rrrrr@{}}\toprule
         Metric & GOMP & J-GOMP & \algname{} $45^{\circ}$ & \algname{} 2G \\
         \midrule
         IE & 330.9 & 121.7 & \textbf{73.6} & 82.5 \\
         IVE & 255.3 & 61.1 & 94.5 & \textbf{18.3} \\
         \bottomrule
    \end{tabular}
    \caption{\textbf{Fragile object transport.} We compute two metrics with respect to the end-effector acceleration vector norm: the integrated error (IE) as the sum of absolute difference from the planned acceleration norm, and the integrated violation of the acceleration constraint (IVE). The end-effector holds a RealSense D435i camera with an IMU to compute the acceleration norm during the motion.}
    \label{tab:imu}
\end{table}

\begin{figure}[t]
    \centering
    \begin{tabular}{@{}c@{\hspace{.05in}}c@{\hspace{.05in}}c@{\hspace{.05in}}c@{\hspace{.05in}}c@{}}
    \includegraphics[width=1.1in,clip,trim=0 0 0 0]{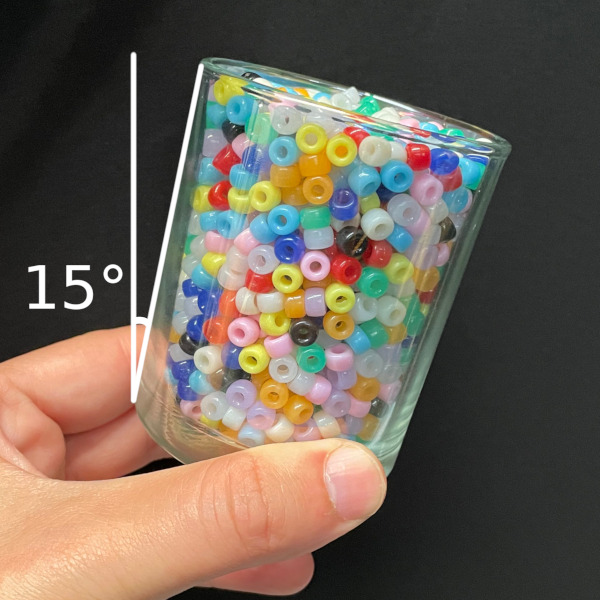} &
    \includegraphics[width=1.1in,clip,trim=0 0 0 0]{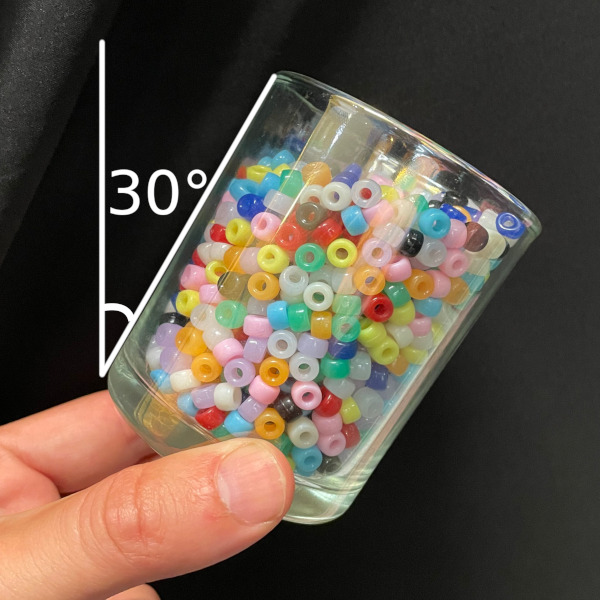} &
    \includegraphics[width=1.1in,clip,trim=0 0 0 0]{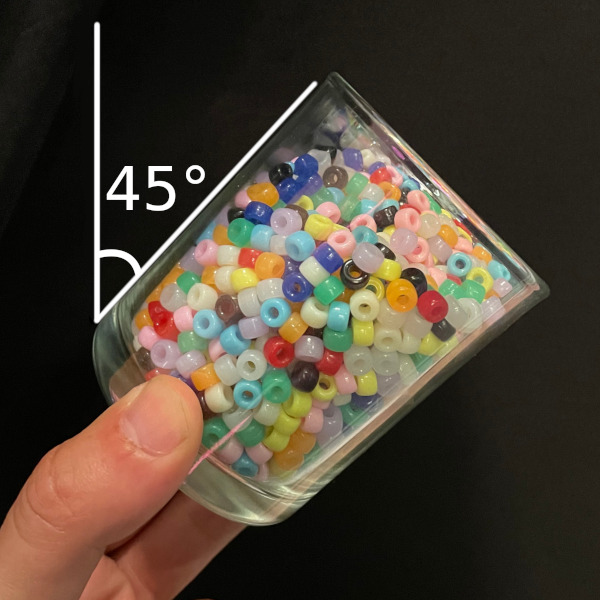} \\
    \end{tabular}
    \caption{\textbf{Fill levels used in open-top container transport experiments.}  As we use beads (instead of liquids) for safety and ease of cleanup, we fill containers to just before spilling.} %
    \label{fig:open_top_fill_levels}
\end{figure}

To test if \algname{} can transport an open-top container without spilling the contents, we task a UR5 robot to transport glass containing beads (Fig.~\ref{fig:over_the_top}).  We vary the fill level based on a tilt-level.  We tilt the container 45$^\circ$, 30$^\circ$, 20$^\circ$, and 15$^\circ$, and fill it so that the beads just barely stay and record the mass (Fig.~\ref{fig:open_top_fill_levels}).  We then have \algname{} compute motions that include the open-top transport constraint (Eqn.~\ref{eqn:alignment}) matching the fill level, and execute the motion.  We vary this for 3 different start-goal pairs and report the results in Tab.~\ref{tab:open_top_lost_mass}.  From the results we observe that \algname{} does not drop a single bead, while all other planners do.  With the (+H) ablations, we observe that running slower is not sufficient for task success, as both baselines spill contents at the same trajectory duration--some of this is attributable to the end-effector taking a longer path to match the trajectory length.  Best performing of the baselines is J-GOMP, which, while consistent with prior observations~\cite{wan2020waiter} that jerk limits reduce spills, limiting jerk alone is insufficient for fast inertial transport.

\begin{figure}[t]
    \centering
    \includegraphics{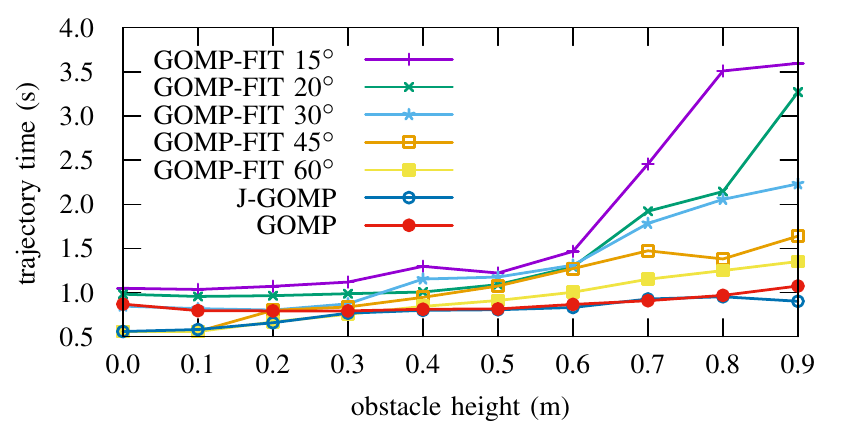}
    \caption{\textbf{Open-top container transport time vs obstacle height.}  \algname{} at various tilt thresholds and baselines compute a transport motion over a wall.  Taller obstacles require longer paths and more time to clear. %
    }
    \label{fig:obstacle_height_vs_trajectory_time}
\end{figure}

We also compare motion time to study how much time is lost due to the open-top constraint, and report the results in Fig.~\ref{fig:obstacle_height_vs_trajectory_time}.  The time loss is surprisingly small, as we observe that \algname{} is able to compute motions that tilt against the inertial forces to keep the contents in the container---most of the time lost is in accelerating and decelerating the object at the beginning and end of the motion.  As obstacle height increases and angle tolerances tighten, the safe passage takes more time to traverse.  With shorter obstacles, GOMP's lack of jerk constraints, requires reducing accelerations limits to prevent protective stops, resulting in worse performance than J-GOMP.  Both J-GOMP and \algname{} can run with higher acceleration limits.

\subsection{Fragile Object Transport}

\begin{figure}
    \centering
    \includegraphics[]{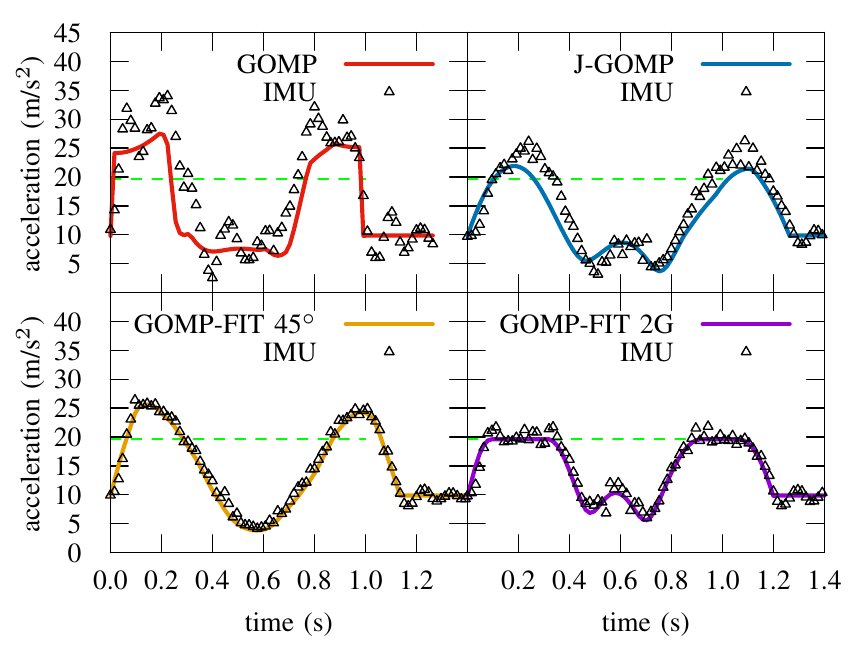}
    \caption{\textbf{Transported IMU readings vs predicted acceleration magnitude} With an end-effector acceleration limit of 2G (green dashed line), GOMP-FIT 2G produces trajectories with IMU readings within a small error, whereas GOMP and J-GOMP have accelerations far exceeding the limit. %
    GOMP-FIT $\ang{45}$ (without acceleration limits) track well with predictions but exceed the limit. %
    }
    \label{fig:imu}
\end{figure}

Transporting a fragile object may require limiting the end-effector acceleration to avoid damaging or dropping the object. To test if \algname{} can reliably limit the end-effector acceleration, we attach a RealSense D435i camera with an Inertial Measurement Unit (IMU) to the UR5 robot end-effector and record the acceleration norm along the executed trajectories. We compute two metrics: the integrated error (IE) as the sum of absolute difference between the end-effector acceleration norm and the planned acceleration norm: 
\[
\sum \left(\lVert A_\mathrm{IMU} - A_\mathrm{traj}\rVert \right),
\]
and the integrated violation error (IVE) as the difference between the end-effector acceleration norm and the acceleration limit:
\[
\sum \max \left(0, \lVert A_\mathrm{IMU} \rVert - \lVert A_\mathrm{max} \rVert \right)
\]
In experiments we set $A_\mathrm{max}$ = 2\text{G} = 19.74\,m/s$^2$ and compare \algname{} with a 2G acceleration limit, with the baselines GOMP, J-GOMP, and \algname{} with a 45$^\circ$-alignment constraint. The results are summarized in Tab.~\ref{tab:imu} and a qualitative result is presented in Fig.~\ref{fig:imu}.  From the measures, we observe that unlike baselines, \algname{} is able to accurately limit accelerations, subject to errors inherent to the sensor and underlying controller.

\subsection{Filled Wineglass Transport}

To test \algname{} for fast inertial transport of a liquid in an open-top container, we compute and execute a motion that transports a glass of wine over a barrier.  Due to results of the experiments with the cup containing beads, and the damage (and stains) that would result from spills, we do not attempt any baseline method.  This experiment is mostly qualitative, as we do not wish to probe the limits of the method in our current lab setup---however, during these experiments, the robot did not spill a drop.  A preview of the accompanying video is in the top-right of Fig.~\ref{fig:motion}.

\section{Conclusion}
\label{sec:conclusion}

We present \algname{}, an optimizing motion planner that solves the fast inertial transport problem of transporting open-top containers, fragile objects, or combinations thereof around obstacles while not spilling, damaging, or losing a grasp due to inertial forces.  \algname{} uses a sequential quadratic program to optimize a discretized trajectory with the first and second derivatives of its configuration, and incorporates non-linear constraints for obstacle avoidance, grasp optimization, and on accelerations at the end-effector.  In experiments on a physical robot, \algname{} was able to reliably and safely transport objects with only minor slowdown compared to motions from fast planners that did not safely transport objects.  Slowed motions of the baseline planners fared little better, suggesting that slowing motions down is not sufficient to achieve reliable inertial transport.

In future work, we will explore addressing the long computation time associated by tuning the optimizer and using deep learning to warm start the computation~\cite{ichnowski2020djgomp}.  The proposed method relies on conservative approximations for joint acceleration limits based on torque limits and transported mass---we hope to tighten up the acceleration limits by transitioning to torque-based limits.  In the current formulation, these would be non-convex constraints that would further slow the computation but result in faster motions.

\section*{ACKNOWLEDGMENT}

This research was performed at the AUTOLAB at UC Berkeley in
affiliation with the Berkeley AI Research (BAIR) Lab, Berkeley Deep
Drive (BDD), the Real-Time Intelligent Secure Execution (RISE) Lab, and
the CITRIS “People and Robots” (CPAR) Initiative. We thank our
colleagues who provided helpful feedback and suggestions. This article solely reflects the
opinions and conclusions of its authors and do not reflect the views of the sponsors or their associated entities.

\bibliographystyle{IEEEtran}
\bibliography{IEEEabrv,references}

\end{document}